\documentclass[conference]{IEEEtran}
\IEEEoverridecommandlockouts
\usepackage{cite}
\usepackage{amsmath,amssymb,amsfonts}
\usepackage{algorithmic}
\usepackage{graphicx}
\usepackage{textcomp}
\usepackage{xcolor}
\usepackage{listings}
\usepackage{array}
\usepackage{hyperref}
\usepackage[doipre={DOI:~}]{uri}

\def\BibTeX{{\rm B\kern-.05em{\sc i\kern-.025em b}\kern-.08em
    T\kern-.1667em\lower.7ex\hbox{E}\kern-.125emX}}
\begin{document}

\title{RabindraNet, Creating Literary Works in the Style of Rabindranath Tagore\\
}

\author{\IEEEauthorblockN{Asadullah Al Galib}
\IEEEauthorblockA{\textit{Department of Computer Science and Engineering, M.Sc in CSE} \\
\textit{BRAC University}\\
Dhaka, Bangladesh \\
asadullah.al.galib@g.bracu.ac.bd}
}

\maketitle

\begin{abstract}
Bengali literature has a rich history of hundreds of years with luminary figures such as Rabindranath Tagore and Kazi Nazrul Islam. However, analytical works involving the most recent advancements in NLP have barely scratched the surface utilizing the enormous volume of the collected works from the writers of the language. In order to bring attention to the analytical study involving the works of Bengali writers and spearhead the text generation endeavours in the style of existing literature, we are introducing RabindraNet, a character level RNN model with stacked-LSTM layers trained on the works of Rabindranath Tagore to produce literary works in his style for multiple genres. We created an extensive dataset as well by compiling the digitized works of Rabindranath Tagore from authentic online sources and published as open source dataset on data science platform Kaggle.
\end{abstract}


\section{Introduction}
The field of NLP, specially the domain of text generation has seen massive improvements in recent years with more sophisticated and powerful models built on top of recurrent neural network models such as LSTM as well as more recent transformer based models. The potential applications of AI generated text in domains such as healthcare and hospitality sector is boundless. However, there is a significant gap in utilizing these advancements in AI to better understand human literature and carry out literary analysis in terms of themes, concepts, societal values, norms at the time of writing. This is more true for Bengali literature due to the lack of quality datasets available online. As a result, our treasured literary works remain completely out of usage for advanced machine learning techniques. \\
In order to bridge this gap between NLP and Bengali literature, we are proposing a character level RNN model based on stacked-LSTM to generate various literary works in the style of Rabindranath Tagore. In the process of creating this model, we also recognised the lack of readily available quality datasets of Tagore that provided granular access to individual literary items such as poems, songs, dramas, essays. To mitigate this issue, we set about creating a compilation of all of Tagore's works in easily accessible formats which can be used for literary analysis as well as inputs for deep learning models.  

\section{Related Work}
Recurrent neural networks, known as RNN are neural networks with hidden state and a recurrence nature which allows them to work with sequential data such as textual data in literature or time series data [3]. However, with very long sequence of textual data, it gets very difficult to train RNNs due to the vanishing or exploding gradient problem, where it gets difficult to pass on gradient backward for training purpose over longer sequence of training data. To solve this problem, various approaches have been applied to effectively train RNNs over longer sequence of data. One of these approaches is to create gated paths for the gradients to travel over long sequence of neural network units without much change. Long short term memory or LSTM and Gated recurrent unit, also known as GRU use this gate mechanism to be effectively trained on datasets with longer sequence. Recurrent neural networks and its variants have been used extensively in text generation due to their ability to handle sequential data most effectively [3]. More recent advancements in text generation have come from auto-encoders and generative adversarial networks [3]. Character level RNN with Hessian-Free optimizer and comparatively small number of hidden units produced high quality text with sophisticated grammatical structure [4]. RNNs have been used to classify poem meters in order to better understand the literary style of the poetry in arabic literature [5]. Recently Transformer based models such as BERT and GPT-2/3 have shown great promise in generating extremely high quality text [6].

\section{Dataset Compilation}
\subsection{Crawling \& Extraction}
One of the main hindrance to advanced literary analysis in Bengali literature using machine learning is the lack of quality datasets in user friendly formats which can be used for multitude of tasks such as exploratory data analysis which demands granular access to individual songs, poems or dramas, at the same time aggregated corpus which can be then fed into deep learning models for prediction or classification tasks. \\
At the start of this project we could not find any open source corpus of Tagore that provided both individual and aggregated content of all of his literary works. To fill in this gap of an open source dataset of the aforementioned type, we initiated a compilation project to collect and prepare a dataset of Tagore that would contain complete collection of works for all of his genres. We found the digitized collection of Tagore's work published online by the Department of Information Technology \& Electronics, Government of West Bengal, India [2] to be an excellent source to compile the target dataset from. Their online collection included entire works of Tagore in the following genres -
\begin{itemize}
    \item Novel
    \item Poem
    \item Song
    \item Story
    \item Essay
    \item Drama
    \item Miscellaneous
\end{itemize}
Each of the genres included collection of individual items. To extract the raw content from the website by maintaining proper hierarchy of items, we created a versatile crawler to handle multiple use cases present in the underlying raw \textit{html} content. The crawler adjusted its extraction methods based on different genres. The entire crawling process includes extraction of collection items, extraction of individual items inside each collection, extraction of paginated content of individual items  and storing data in raw textual format after parsing from html (see Figure 1).
\begin{figure}
\includegraphics[width=0.7\linewidth]{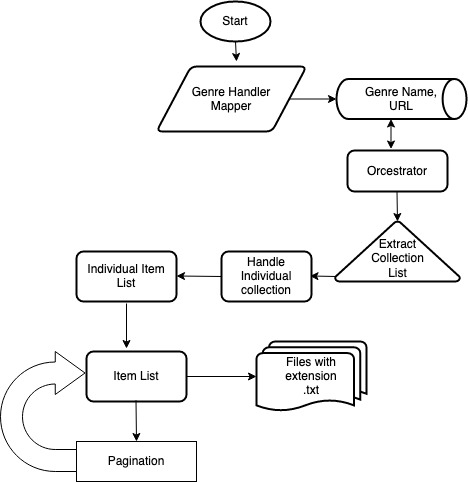}
\centering
\caption{Crawling flow chart} \label{fig1}
\end{figure}

\subsection{Preprocessing \& Formatting}
After collecting individual literary items in raw textual format from each genre, we placed them in hierarchy by maintaining proper genre and parent collection. After that, we executed an aggregator that compiled all individual items of a genre into a single file. Before adding each item into corresponding genre, we passed raw text through a basic pre-processor which removed title names, page numbers and other special symbols using regex rules. We prepared two file formats for each genre, \textit{CSV} and \textit{TXT}. The \textit{CSV} format contained comma separated values including literary item name, parent collection, genre and content. This type of format is useful for various exploratory analysis. On the other hand \textit{TXT} format contained aggregated content from all individual literary items concatenated together which is useful for training deep learning models. We then published this dataset in the data science competition platform Kaggle for the benefit of the community [1].

\section{Methodology}
We used the prepared dataset from previous step [1] to train a character level RNN model consisting of single LSTM layer. For initial step, we trained the model only on poem dataset with epoch set to 40. After training the model, we studied the quality of the generated text for poem genre. See Figure 2 for an excerpt of the generated poem,
\begin{figure}
\includegraphics[width=0.5\linewidth]{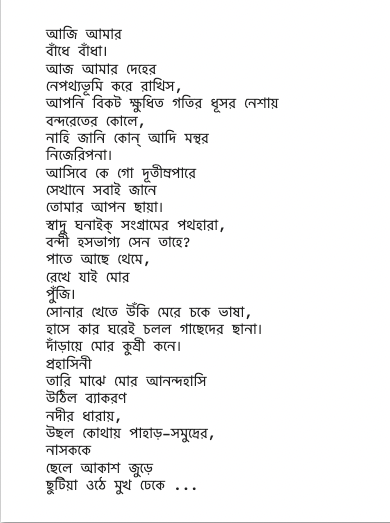}
\centering
\caption{Generated poem from single LSTM model} \label{fig2}
\end{figure}

From the generated text it is evident that the model learned basic word formation, however it failed to learn how to choose consecutive words so that it would form a coherent content. From this finding, we were inspired to explore stacked-LSTM in order to increase the model capacity for abstract understanding over long sequence of data and give the model ability to learn word association better.\\
Different sections of the process starting from analyzing the dataset to generating new content are described below in details (see Figure 3). In our work, we experimented with datasets from 3 different genres, namely song, poem and story. All the dataset samples provided in the following sections are taken from song genre.
\begin{figure}
\includegraphics[width=0.5\linewidth]{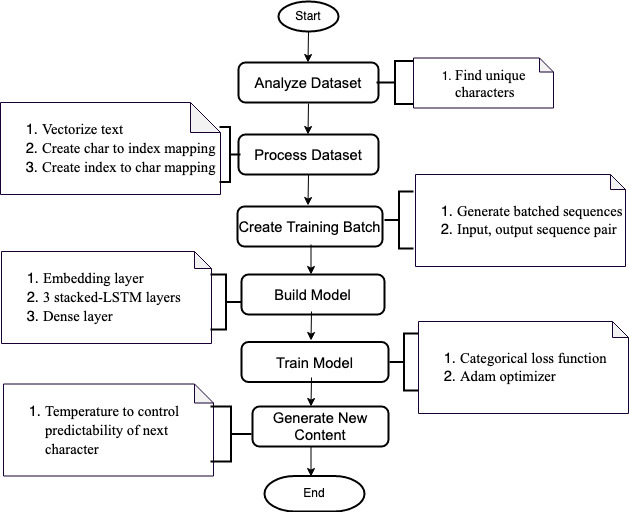}
\centering
\caption{Methodology diagram} \label{fig3}
\end{figure}

\subsection{Analyze Dataset}
At first we analyzed the dataset to find out basic information regarding the dataset at hand. The song dataset had total \textit{733280} characters. Following is an excerpt form the dataset (see Figure 4).
\begin{figure}
\includegraphics[width=0.5\linewidth]{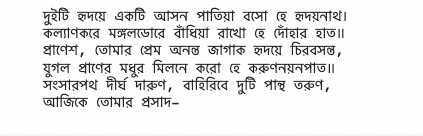}
\centering
\caption{Excerpt from song dataset} \label{fig4}
\end{figure}

The dataset had \textit{100} unique characters. Some of which are listed below (see Figure 5).
\begin{figure}
\includegraphics[width=0.5\linewidth]{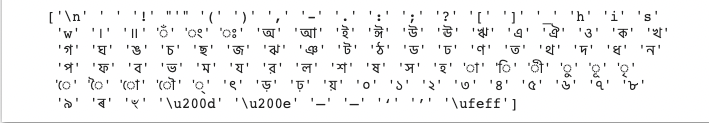}
\centering
\caption{Unique characters} \label{fig5}
\end{figure}

\subsection{Process Dataset}
At this stage, the dataset consisted of sequence of characters. In order to feed the data into our training model we had to convert the dataset from a sequence of characters to a sequence of numbers. To achieve that, we first created a vocabulary or mapping of character-to-index using the unique character extracted in the previous step and replaced all the characters in the dataset with corresponding index numbers. Followings are some samples from the character to index mapping. 

\begin{lstlisting}
  "'" :   3,
  '(' :   4,
  ')' :   5,
  ',' :   6,
  '-' :   7,
  '.' :   8,
  ':' :   9,
  ';' :  10,
  '?' :  11,
  '[' :  12,
  ']' :  13,
  ...
\end{lstlisting}
We also created an index-to-character mapping to be used during the text generation step to convert the predicted index number to corresponding character.

\subsection{Create Training Batch}
To create training batch, we first divided the dataset into sequences of \textit{101} characters. For our experiment, we considered sequence length as \textit{100}. The additional character in the sequence was used during the shifting of each sequence to generate input and corresponding output sequence. After this step, the dataset had altogether \textit{7260} sequences of length \textit{101} for each epoch. After that, we created input-output sequences from each sequence in the dataset by removing the trailing index for the input sequence and first index for the output sequence. The result is \textit{7260} input and output sequence pairs of length \textit{100}. See Figure 6 for an example.
\begin{figure}
\includegraphics[width=0.7\linewidth]{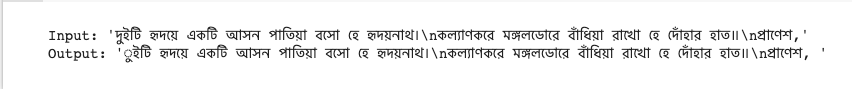}
\centering
\caption{Input output sequence} \label{fig6}
\end{figure}

During training, at the first step, the model receives the index number for first character as the input and using that it tries to predict the index number for the next character. During next step, model receives the index number of the next character and uses that and previous step context to predict the index for next character (see Figure 7). 
\begin{figure}
\includegraphics[width=0.7\linewidth]{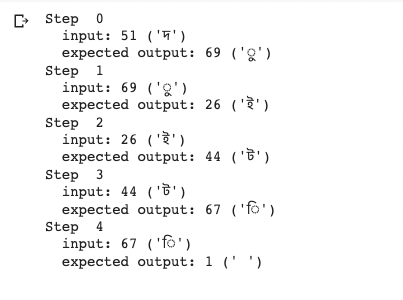}
\centering
\caption{Next character prediction using previous character} \label{fig7}
\end{figure}
To run the training process efficiently, we shuffled the dataset and divided into batch size of \textit{64}. 
\subsection{Build Model}
Finally, we built the model using \textit{Embedding layer} as our first input layer, then we stacked 3 \textit{LSTM layers} as our RNN layers before adding a \textit{Dense layer} as the final and output layer.
We set the embedding dimension of the first layer to \textit{256} and used \textit{1024} RNN units for each LSTM layer. The output dimension was the vocabulary size which was \textit{100} for the song dataset. Followings are the model summary (see Figure 8) and structure (see Figure 9),

\begin{figure}
\includegraphics[width=\linewidth]{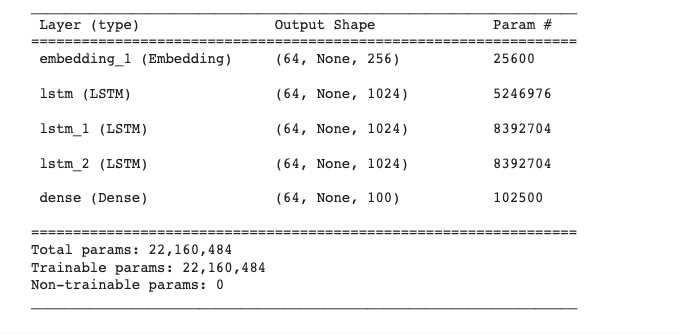}
\centering
\caption{Model summary} \label{fig8}
\end{figure}

\begin{figure}
\includegraphics[width=0.7\linewidth]{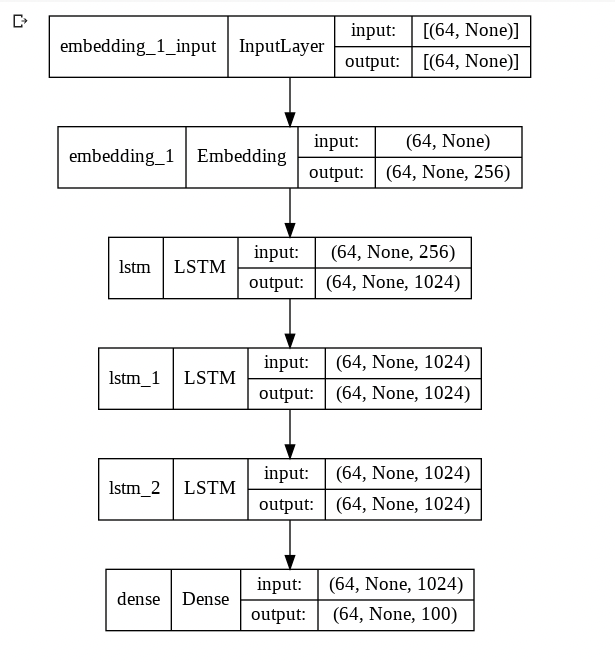}
\centering
\caption{Model structure} \label{fig9}
\end{figure}

For each character, the model passes it into the embedding layer to get the embedding of stated dimension and executes the LSTM for one step to generate the log likelihood of the next character using the output dimension of vocabulary length. Before training, the model does not have any knowledge of the dataset. See Figure 10 for a sample of generated using initial parameters of the model before the training step.
\begin{figure}
\includegraphics[width=0.7\linewidth]{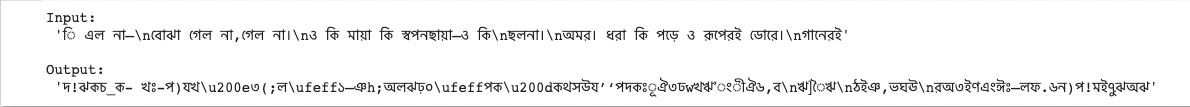}
\centering
\caption{Predicted output before training} \label{fig10}
\end{figure}

\subsection{Train Model}
We treated the training stage similar to a classification problem. Given the state of the model up to previous time step and current input character, the model had to predict the class of next character which had the dimension of \textit{100}, the vocabulary size. To train the model, we used \textit{Adam optimizer} and \textit{Categorical cross entropy} as the loss function and ran it for 40 epochs for all datasets. 

\subsection{Generate New Content}
After the model had been trained, we generated new text by providing it with a starting sequence of characters and the number of characters to generate. We also used a temperature variable to control the predictability of next character selection. For each generated character, we use it as the previous character and feed it into the model to generate the next one.

\section{Results}
As previously mentioned, we executed the same training procedure with 3 different datasets from different genres, namely song, poem and story. First we present the output of the model after it was trained on poem dataset for single-layer LSTM (see Figure 11) and 3-layer stacked-LSTM (see Figure 12) with temperature value set to 1 and starting text as \textbf{"Aji Amar"}\\
We can observe that stacked LSTM model not only learned how to form correct words but the generated text seems more coherent than the single layer LSTM model. However when we provided the stacked-LSTM model with same starting sequence of characters but temperature value of 1.5, it produced more surprising text (see Figure 13).\\

Next, we present the output of the stacked-LSTM model after it was trained on song dataset with different temperature value (see Figure 14, 15).\\
We can notice the 3-fold increase in elapsed time during training from single-layer LSTM to stacked-LSTM with epoch=40,\\
\textbf{Elapsed time for training the single later LSTM with epoch=40:}
\begin{itemize}
    \item Poem: 3113s
    \item Story: 3095s
\end{itemize}
\textbf{Elapsed time for training the 3-Layer stacked LSTM with epoch=40:}
\begin{itemize}
    \item Poem: 9934s
    \item Story: 10175s
\end{itemize}

\begin{figure}
\includegraphics[width=0.7\linewidth]{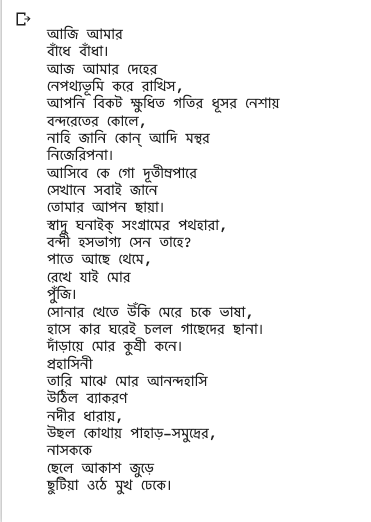}
\centering
\caption{Generated poem from single layer LSTM} \label{fig11}
\end{figure}

\begin{figure}
\includegraphics[width=0.7\linewidth]{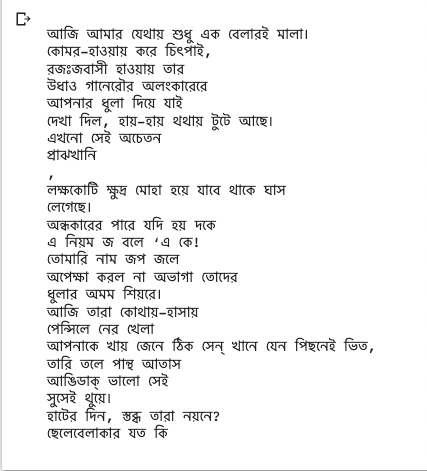}
\centering
\caption{Generated poem from 3 layer stacked-LSTM} \label{fig12}
\end{figure}

\begin{figure}
\includegraphics[width=0.7\linewidth]{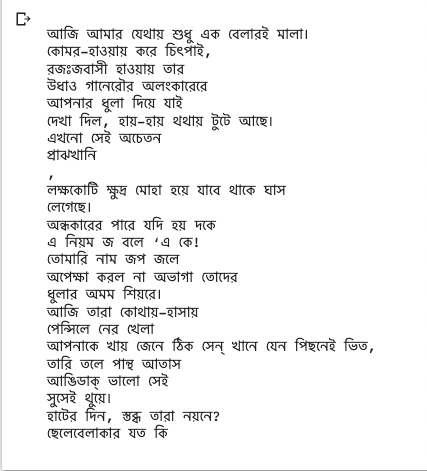}
\centering
\caption{Generated poem from 3 layer stacked-LSTM, temperature=1.5} \label{fig13}
\end{figure}

\begin{figure}
\includegraphics[width=0.7\linewidth]{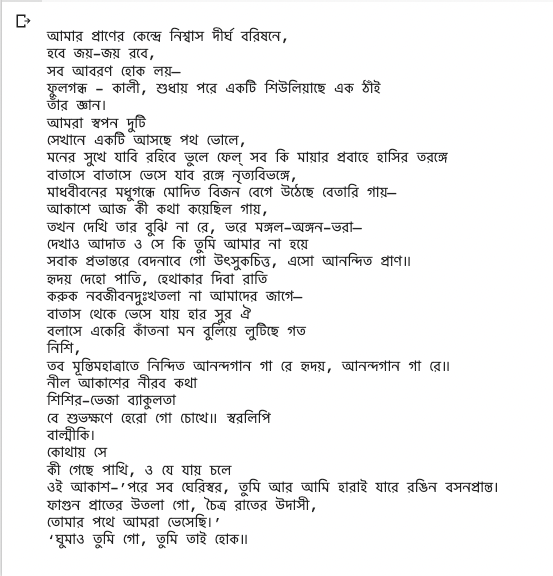}
\centering
\caption{Generated song from stacked-LSTM, temperature=1.0} \label{fig14}
\end{figure}

\begin{figure}
\includegraphics[width=0.7\linewidth]{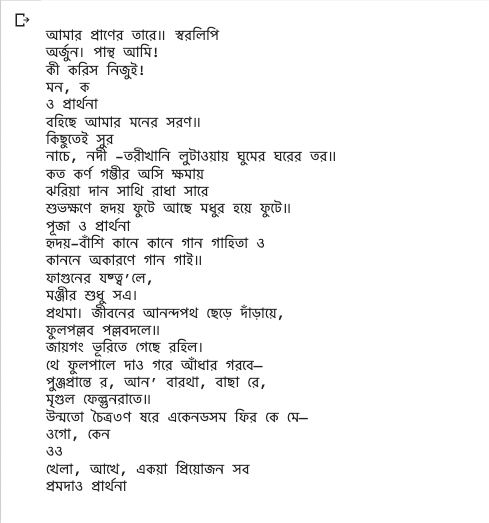}
\centering
\caption{Generated song from stacked-LSTM, temperature=1.5} \label{fig15}
\end{figure}

\section{Future Works}
In future, we wish to train the stacked-LSTM model for larger epoch values and evaluate the generated text with evaluation metrics other than human perception using literary analysis. We also hope to experiment with other generative models and compare the results with the current LSTM model. Finally, we are working towards publishing this model for general public to explore and enjoy by generating texts using trained models for different genres. Currently we are producing texts of \textit{500} to \textit{1000} characters of three genres which the model was trained upon. We plan to experiment with much longer texts as found in stories and essays.

\section{Conclusion}
We have taken the first step of diving into Bengali literature and experimenting with text generation in the style of the literary giant Rabindranath Tagore using character level RNN model with stacked-LSTM layers. We have shown that simple stacked LSTM layers can produce high quality text in literary genres such as poem, song and story. This is promising to take on the challenges of much longer texts involving dramas or essays. Gradually we hope to work on more advanced theme analysis and comparative analysis of literary styles of Bengali writers using more sophisticated NLP tools.


\end{document}